\documentclass[10pt,twocolumn,letterpaper]{article}

\usepackage[pagenumbers]{cvpr}

\AtBeginDocument{%
  \providecommand\BibTeX{{%
    \normalfont B\kern-0.5em{\scshape i\kern-0.25em b}\kern-0.8em\TeX}}}

\usepackage{listings}
\usepackage{xcolor}

\usepackage{graphicx}
\usepackage{amsmath}
\usepackage{booktabs}

\usepackage{tabu}
\usepackage{multirow}
\usepackage{rotating} %
\usepackage{caption}
\usepackage{subcaption}
\usepackage{pifont}

\usepackage{soul}
\usepackage{svg}

\DeclareMathAlphabet{\altmathcal}{OMS}{cmsy}{m}{n}
\DeclareMathAlphabet{\mathbfit}{OT1}{ptm}{bx}{it}

\newlength\paramargin
\newlength\figmargin

\newlength\secmargin
\newlength\figcapmargin
\newlength\tabcapmargin

\newcommand{\myparagraph}[1]
{
\vspace{2mm}\noindent\textbf{#1}
}

\long\def\ignorethis#1{}

\newcommand {\jiabin}[1]{}
\newcommand {\junyan}[1]{}

\definecolor{methodred}{RGB}{248, 203, 173}
\definecolor{methodyellow}{RGB}{245, 220, 143}
\definecolor{methodgreen}{RGB}{197, 224, 180}
\definecolor{methodblue}{RGB}{180, 199, 231}

\newcommand{\tb}[1]{\textbf{#1}}

\newbox\jsavebox%

\makeatletter
\newcommand{\providelength}[1]{%
  \@ifundefined{\expandafter\@gobble\string#1}
   {%
    \typeout{\string\providelength: making new length \string#1}%
    \newlength{#1}%
   }
   {%
    \sdaau@checkforlength{#1}%
   }%
}

\newcommand{\sdaau@checkforlength}[1]{%
  \edef\sdaau@temp{\expandafter\sdaau@getfive\meaning#1TTTTT$}%
  \ifx\sdaau@temp\sdaau@skipstring
    \typeout{\string\providelength: \string#1 already a length}%
  \else
    \@latex@error
      {\string#1 illegal in \string\providelength}
      {\string#1 is defined, but not with \string\newlength}%
  \fi
}
\def\sdaau@getfive#1#2#3#4#5#6${#1#2#3#4#5}
\edef\sdaau@skipstring{\string\skip}
\makeatother

\def\xi{\mathbf{x}_i}

\graphicspath{{figures}, {examples}}

\newcommand{\cmark}{\checkmark}%
\newcommand{\xmark}{$\times$}%

\def\confName{CVPR}
\def\confYear{2023}
\begin{document}
\title{Coherent Zero-Shot Visual Instruction Generation}

\author{Quynh Phung~~~~Songwei Ge~~~~Jia-Bin Huang\\
    University of Maryland College Park\\
\url{https://instruct-vis-zero.github.io}
}

\newcommand{\Sref}[1]{Sec.~\ref{#1}}
\newcommand{\Eref}[1]{Eq.~(\ref{#1})}
\newcommand{\Fref}[1]{Fig.~\ref{#1}}
\newcommand{\Tref}[1]{Table~\ref{#1}}
\newcommand{\R}[0]{\mathbb{R}}

\twocolumn[{%
\renewcommand\twocolumn[1][]{#1}%
\maketitle
\vspace{-3em}
\begin{center}
    \centering
    \captionsetup{type=figure}
    
    \includegraphics[width=1.\textwidth]{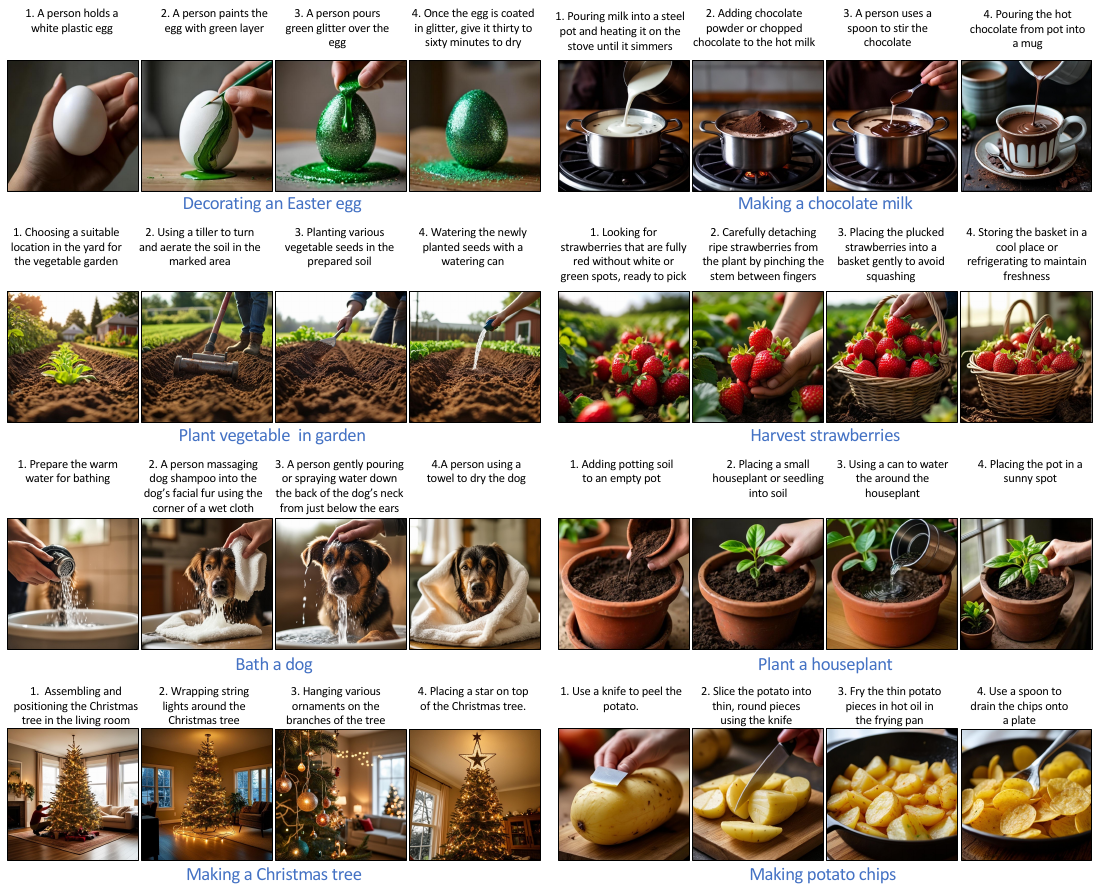}
  
   \setcounter{figure}{0}
   \captionof{figure}{
      \tb{Visual instruction generation results.}
      Given a sequence of textual instructions for a certain task, our method generates the \emph{visual} instructions that illustrate the individual steps.
      Our method is training-free and thus preserves the quality and generalizability of the underlying image generation models. 
      We showcase the generated visual instructions for different tasks from cooking to gardening. The samples possess high visual quality, align with the instructions, and maintain coherent object identity with desired changes at each step.
      }
  \label{fig:teaser}
\end{center}%
}]
    
\begin{abstract}
  Despite the advances in text-to-image synthesis, particularly with diffusion models, generating visual instructions that require consistent representation and smooth state transitions of objects across sequential steps remains a formidable challenge. 
  This paper introduces a simple, training-free framework to tackle the issues, capitalizing on the advancements in diffusion models and large language models (LLMs). 
  Our approach systematically integrates text comprehension and image generation to ensure visual instructions are visually appealing and maintain consistency and accuracy throughout the instruction sequence. 
  We validate the effectiveness by testing multi-step instructions and comparing the text alignment and consistency with several baselines.
  Our experiments show that our approach can visualize coherent and visually pleasing instructions. 
\end{abstract}

\begin{figure}[t]
  \centering
    \includegraphics[width=1.0\columnwidth]{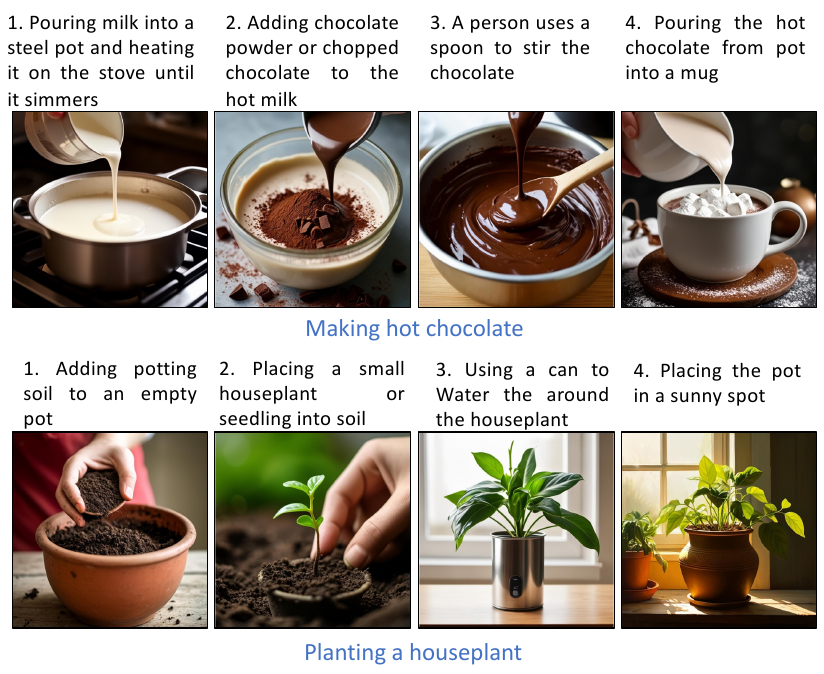}
  \caption{\tb{Limitation of text-to-image generation in visual instructions task .}
  The crucial components of good visual instruction are 1) alignment with the text-based instruction and 2) coherence across different steps demonstrating the state changes. 
  The current text-to-image generation methods focus only on the former. 
  Consequently, the results may confuse the readers.
  In this paper, we develop a training-free method to enable a more coherent visual instruction generation.
  }
  \label{fig:motivation}
\end{figure}

\section{Introduction}
\label{sec:introduction}

Textual instructions are among the most prevalent tools for grasping new skills and knowledge and solving real-world problems, large and small. 
Generating visual illustrations of instructions has been a vital research problem~\cite{du2024learning,yang2024learning,hong2022cogvideo} for their straightforwardness and ability to transcend language barriers, providing an intuitive understanding of the textual instructions~\cite{yadav2011if}. 
In addition, apart from conveying information to humans, such visual data has also been widely adopted to train robotic policies~\cite{ebert2021bridge,brohan2022rt}. 

In this paper, we focus on generating \textit{static} visual instructions. 
Unlike generating instructional videos~\cite{miech2019howto100m,zhou2018towards} that demand temporal consistency, generating static visual illustrations poses a relatively more approachable problem~\cite{genhowto}. 
We leverage the recent advancements in text-to-image diffusion models~\cite{podell2023sdxl,ho2022imagen,balaji2022ediffi}, which has shown remarkable zero-shot capacity and photorealism, to visualize the instructions across a wide range of problem categories.
Instead of fine-tuning the model on the instructional image dataset as in the existing methods~\cite{genhowto,menon2023generating,bordalo2024generating}, which could compromise the generation quality and limit itself to certain categories, we develop a training-free method of generating visual instructions.

Directly inputting the instructions to the text-to-image models incurs several issues, as shown by the examples in Figure~\ref{fig:motivation}. First, most instructions explain the procedure of actions, while the text-to-image model expects the description of the image content. As shown in step 2 of the ``Making hot chocolate'' example, the milk container becomes a glass bowl instead of the steel pot as mentioned in step 1 due to the lack of information in the instruction. 
This necessitates an approach to bridging the gap between instructional texts and the conditioning text used for image generation. 
Given the procedures at the current and previous steps, one needs to infer the proper image content at the current step.
To this end, we propose an instruction re-captioning strategy~\cite{betker2023improving,omost} to convert the instructional texts into actions and states using large language models (LLMs). We show that combining the action and state as the condition significantly enhances the quality and relevance of the generated illustrations.

In addition, the objects' identities may alter arbitrarily across different steps. 
For example, in the "planting a houseplan" example, the shape and texture of the pot in the first step are different from those in steps 3 and 4. 
This poses a common challenge when using text-to-image models to generate multiple images - there is often a lack of coherency across the generated illustrations. 
Recent studies have made progress in maintaining consistency for human portraits~\cite{chen2024idaligner,guo2024pulid,huang2024consistentid}, where the identity features can be derived from the models trained on extensive human-centric datasets. However, our problem cannot benefit from the same idea since the identity features of the general object categories, such as kitchen utensils, are unavailable. 
Therefore, we refer to the general methods to achieve consistency, such as feature sharing and injection~\cite{tewel2024training,tumanyan2022plug}.

Although these methods improve the generation consistency, they also induce the ``over-consistent'' problem in visual instruction generation. Specifically, many instructions involve changed objects among different steps. For instance, a recipe may contain a raw ingredient that is chopped, seasoned, and presented in entirely different states throughout the process. 
This inherent variability complicates the problem of maintaining object consistency and makes it unsuitable to use the vanilla feature-sharing strategy. 
To this end, we propose an adaptive feature-sharing method with finer-grained constraints.
First, similar to the previous method, we adopt a local region constraint to enforce sharing to happen on certain pixels. 
However, localizing the objects using attention maps becomes less reliable when the base model architecture differs from the UNet in Stable Diffusion~\cite{cascade}. Instead, we utilize large-scale segmentation models to produce the masks~\cite{li2023semantic,kirillov2023segany,liu2023grounding,ren2024grounded,huynh2023simpson,huynh2024maggie}. 
Second, we apply a global constraint on the feature-sharing scale between each pair of steps based on their similarity. We exploit the reasoning capabilities of the LLMs to develop a similarity matrix that characterizes such state similarity. We show that our adaptive feature-sharing method enables the variability of objects across instructional steps while preserving object identity.

As showcased in Figure~\ref{fig:teaser}, our proposed method can generate high-quality, consistent visual representations based on instructional texts.
We also perform quantitative experiments with various evaluation metrics to ablate individual components of our method on the text-image alignment and consistency.
However, we notice that the image similarity metric only emphasizes the consistency aspect and conflicts to achieve variation across different steps. To this end, we propose a framework to evaluate the visual instruction generation quality using large-scale visual language models. 
We will release the code and full evaluation suite for reproducible research. The contributions of our work can be summarized as follows:
\begin{enumerate}
    \item We develop a \textit{training-free} method for generating visual instructions with pre-trained text-to-image diffusion models.
    \item We propose an illustration re-captioning strategy, greatly improving the generation quality and relevance.
    \item We introduce an adaptive feature-sharing method with finer-grained constraints to maintain object identity across different steps while allowing for necessary variations.
    \item We present a framework to evaluate the visual instruction quality using large-scale visual language models. We show that our method can preserve the generation quality and show applicability across various categories.
\end{enumerate}

\section{Related work}
\label{sec:relatedwork}

\myparagraph{Text-to-Image Generation with Diffusion Models}. 
Diffusion models~\cite{ho2020denoising,song2021denoising,song2021scorebased,song2019generative} have become the ubiquitous choice for visual generation for their effectiveness in scaling on visual data distribution. 
When training on the large-scale datasets~\cite{kakaobrain2022coyo-700m,schuhmann2022laionb}, improved training and sampling with advanced techniques~\cite{ho2022cascaded,rombach2022high,nichol2021improved,ho2022classifier}, the state-of-the-art results have been achieved in image~\cite{ramesh2022hierarchical,saharia2022photorealistic,balaji2022ediffi,dai2023emu,podell2023sdxl,pernias2024wrstchen} and video~\cite{ho2022video,singer2022make,girdhar2023emu,ge2023preserve,bar2024lumiere} generation. 
As learning from billions of data samples, the pretrained diffusion models have been shown to have great generalization capacity and thus been adapted to various downstream applications such as image editing~\cite{meng2022sdedit,hertz2022prompt,parmar2023zero}, controllable generation~\cite{zhang2023adding}, 
personalized image generation~\cite{ruiz2022dreambooth,kumari2022multi}, and even non-generative tasks~\cite{tang2024emergent,li2023your}. In this paper, we explore applying these pre-trained models to generate visual guidance in a zero-shot way.

\myparagraph{Improving Consistency in Image Generation}.
Generating consistent images has been an important sub-problem in different tasks, including video generation~\cite{tokenflow2023,text2video-zero,wu2023tune,Zhou2024storydiffusion}, multi-view image generation~\cite{shi2024mvdream,liu2024syncdreamer,tseng2023consistent}, and character generation~\cite{wang2024characterfactory,chen2024idaligner,guo2024pulid,huang2024consistentid,tewel2024training}. 
Compared with these tasks, we focus on improving the consistency of the shared objects in different steps of the visual instructions. 
Unlike video or multi-view image generation, we don't enforce hard-constraint geometrical or temporal consistency. 
Different from the character generation that only cares about human identity, we need to deal with general objects that don't have ID feature extractors as those for humans~\cite{wang2024characterfactory,chen2024idaligner,guo2024pulid,huang2024consistentid}.

To achieve better consistency, many existing studies resort to fine-tuning~\cite{ge2023preserve,Liu_2023_ICCV,huang2024consistentid} partially or completely the diffusion models on the consistent image data. In this paper, we tackle with zero-shot consistent image generation~\cite{tewel2024training}. 
Previous studies have found that features in the diffusion models encode different information and can be utilized to control the generation. Cross-attention maps connect generation with text prompts and can be manipulated for additional textual information~\cite{phung2023grounded,chefer2023attend,hertz2022prompt,balaji2022ediffi}. 
Self-attention maps link pixels and encode rich structural information, which has been utilized to extract or modify the layouts~\cite{phung2023grounded,cao2023masactrl,patashnik2023localizing}. 
The feature maps and noised latents contain more detailed information and can be used to reproduce the exact intact regions like background~\cite{tumanyan2022plug,ge2023expressive}. 
In this paper, we build on these observations and propose several techniques to improve the consistency required in visual instruction generation. 
Specifically, we combine the controllability offered by these methods with the semantical understanding capacity of Large Language Models~\cite{lian2024llmgrounded,feng2024layoutgpt} for finer-grained coherency.

\myparagraph{Visual Instruction Generation}.
Both video and image can serve as the media for visual instructions. 
We focus only on generating static images as visual instructions~\cite{genhowto,menon2023generating,bordalo2024generating}. 
Earlier works on instruction generation include recipe generation~\cite{salvador2017learning,salvador2019inverse,Chhikara_2024_WACV}, while we are also interested in illustrating recipe steps with visual instructions. 
More recent works leverage the great generative power of pre-trained diffusion models and fine-tune these models on the visual instruction datasets~\cite{miech2019howto100m,yang2021visual}. 
Bordalo et al.~\cite{bordalo2024generating} integrates an Alpaca-7B model with a Stable Diffusion model for fine-tuning and generating sequences of visual illustration of recipes. 
GenHowTo~\cite{genhowto} curates a dataset of states, actions, and resulting transformations triplets and trains a conditioned diffusion model on it. 
StackedDiffusion~\cite{menon2023generating} fine-tuned a pre-trained text-to-image diffusion model with the stacked input on the Visual GoalStep Inference (VGSI) dataset~\cite{yang2021visual}. 
Different from these existing methods, we aim to use the \emph{pre-trained} diffusion model in a \emph{zero-shot} manner for visual illustration generation.

\section{Method}

\begin{figure*}[t]
    \centering
    \begin{subfigure}[a]{1.\textwidth}

    \includegraphics[width=1.\textwidth]{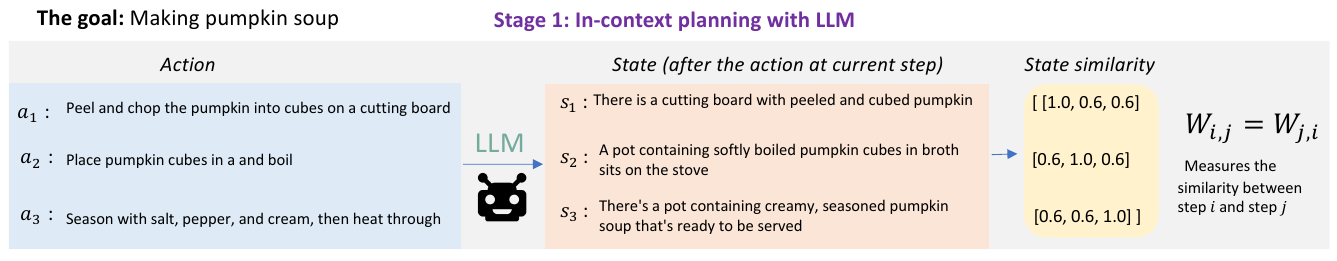}
    \end{subfigure}
    \begin{subfigure}[b]{1\textwidth}
    \includegraphics[width=1.\textwidth]{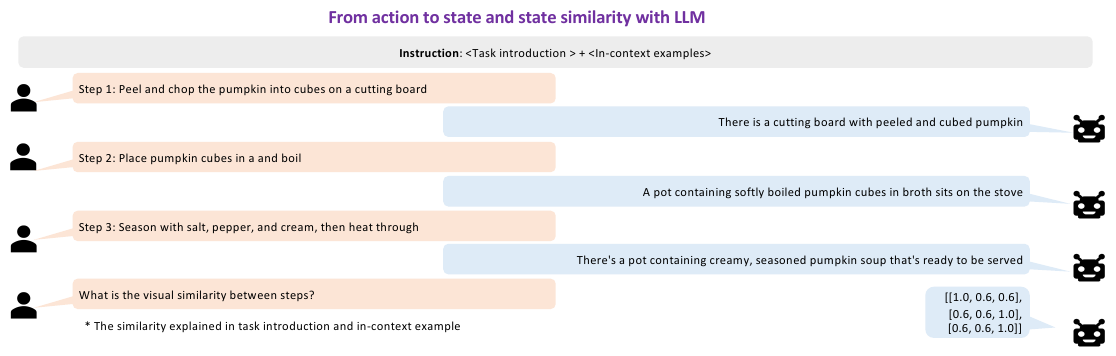}
    \end{subfigure}
    \begin{subfigure}[b]{1\textwidth}
    \includegraphics[width=1.\textwidth]{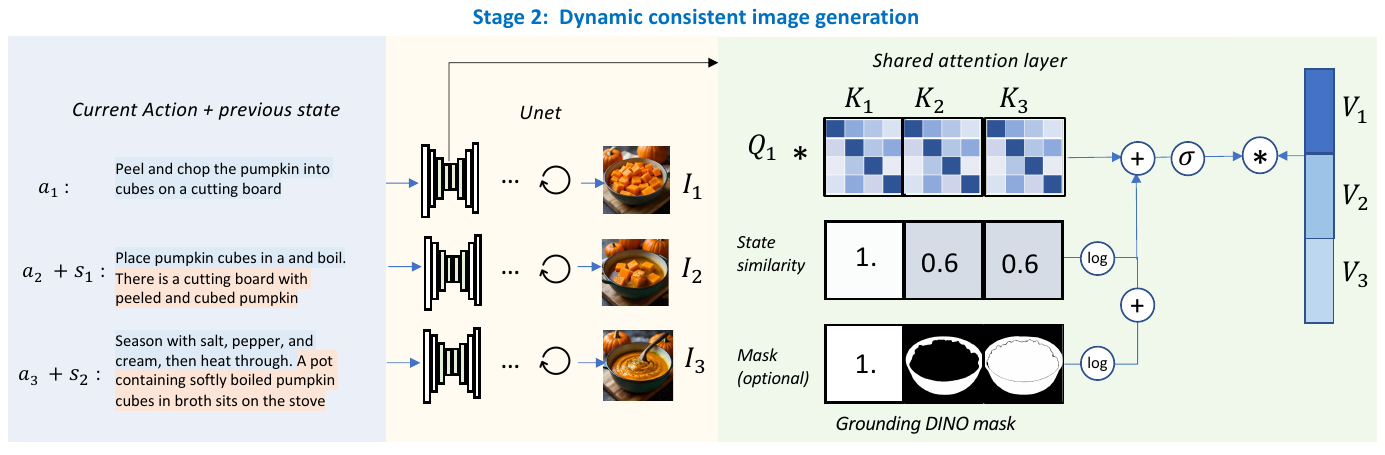}
    \end{subfigure}
    \caption{
    \tb{Our framework for zero-shot instruction visualization.}
    Our framework operates in two distinct phases. 
    In the first phase, we use an LLM (e.g., the GPT-4 model) to generate the scene state after each step in the list of instructions. 
    The generated scene state helps guide the image generation in the next stage.
    We also ask the LLM to generate the similarity between states.
    This matrix, with each row indicating the visual similarity of a current visual step to others, guides the generation process. 
    For example, to achieve high state similarity, we wish to maintain consistency as much as possible across the two steps. 
    A low state similarity indicates the performed action changes the scene state substantially. 
    In such cases, blindly encouraging \emph{consistency} across steps may hurt the quality of the visualized instruction image. 
    In the second phase, we utilize a shared attention layer—replacing the standard model—to allow queries from one image to access keys and values from others within the same instruction set. 
    We enhance this sharing mechanism by applying standard attention masking, controlled by the similarity matrix, to finely tune the interaction between visual elements.
       }
    \label{fig:overview}
\end{figure*}

Given a set of instructions, we harness a pre-trained text-to-image diffusion model to generate the visual illustrations.
As shown in~\Fref{fig:overview}, our approach contains two major stages. 
First, to fill the distribution gap between the instructions and image descriptions, we perform \emph{in-context planning} with LLMs to re-caption the instructions. 
Second, given the re-captioned instructions, we propose an adaptive feature-sharing method for dynamic, coherent image generation. 
In both stages, we use off-the-shelf pre-trained models \emph{without any extra training}. 

\subsection{Preliminaries}
\myparagraph{Text-to-image diffusion models.}
The text-to-image diffusion model incorporates a denoiser network $D$ that is trained to estimate the noise in the current image, \(\mathbf{\epsilon}_t = D(\mathbf{x}_t; \mathbf{c})\), where \(t\) represents the timestep, and \(\mathbf{c}\) denotes the conditional information embedding. 
During the inference time, an initial random Gaussian noise is iteratively denoised to generate a real image.

\myparagraph{Self-attention layer} is essential in $D$ for integrating global information across the entire image. 
It redistributes the features from each spatial location to similar regions.
Suppose that $\mathbf{x} \in \mathbb{R}^{w \times h \times d}$ denote the input feature map of some self-attention layer, where $w$, $h$, and $d$ are the width, height, and dimension. 
Let \( P = h \times w \) for simplicity. 
By applying linear mappings to the feature map $\mathbf{x}$ to obtain the \emph{key} $K \in \mathbb{R}^{P \times d_k}$, \emph{value} $V \in \mathbb{R}^{P \times d_v}$, and \emph{query} $Q \in \mathbb{R}^{P \times d_k}$ where $d_k$ is dimension of key $K$ and query $Q$, $d_v$ is dimension of value $V$, the self-attention map $A^t$ at step $t$ is generated by :
\begin{equation}
\label{eq:attention}
    A^t = \text{softmax} \left(\frac{QK^\top}{\sqrt{d_k}}\right) \in [0,1]^{P\times P},
\end{equation}
The attention maps mechanism calculates the similarity between query (Q) and key (K), which determines how much attention each value (V) receives.

\myparagraph{KV sharing.} To maintain object consistency across institutional steps, we draw inspiration from techniques used in video generation~\cite{text2video,wu2023tune}. These studies share keys and values within the self-attention layers across frames to allow the queries to attend to consistent elements in previous frames, implemented by concatenating keys and values:
\begin{equation*}
K^+ = [K_1 \oplus K_2 \oplus \dots \oplus K_N], V^+ = [V_1 \oplus V_2 \oplus \dots \oplus V_N]
\end{equation*}
\begin{equation}
\label{eq:attention_maps}
A_i^+ = \text{softmax}\left( \frac{Q_i K^{+T}}{\sqrt{d_k}} \right) \in [0,1]^{P \times N \cdot P}  
\end{equation}
\begin{equation}
H_i = A_i^+ \cdot V^+ \in \mathbb{R}^{P \times d_v},
\end{equation}
where$ \oplus $ denotes concatenation, $N$ is the number of images, and $i\in \{1, 2, \cdots N\}$. However, this technique may not generalize to our problem because the generated frames are supposed to be similar in both background and foreground, where visual instructions may contain dynamic elements.
 
Consistory~\cite{consistory} further proposed Self-Driven Self-Attention, focusing on sharing keys and values within individual objects across different frames. 
Specifically, it extracts object masks $M$ from the cross-attention maps and assigns a $-\infty$ score to the self-attention maps for any pixel where the object mask value is zero, indicating that there should be no sharing in these regions. 
This updates the self-attention map calculation with the object masks:
\begin{equation}
M_i^+ = [M_1 \dots M_{i-1} \oplus M_i \oplus M_{i+1} \dots M_N],
\end{equation}
\begin{equation}
\label{eq:attention_maps}
A_i^+ = \text{softmax}\left( \frac{Q_i K^{+T}}{\sqrt{d_k}} + \log M_i^+ \right) \in \mathbb{R}^{P \times N \cdot P},
\end{equation}
where $M_i=\mathbf{1}$, corresponding to the images $i$-th. This ensures that each image only attends to itself or object regions of other images. However, this method still assumes the objects to be fully consistent across different images, which is not often the case in visual instructions. 

\subsection{Re-captioning instructions as descriptive texts}
Generating visual instructions from a sequence of textual instructions presents a significant challenge to current text-to-image models. 
To solve the problem, the model needs to understand objects' states and relationships across successive steps.
For instance, as shown in the \Fref{fig:motivation}, consider the first two steps of making hot chocolate, where the milk is first poured into a steel pot, and the chocolate is then added to the milk. If one directly uses instructions as the input, the text-to-image models will not be informed with the context of ``milk in the steel pot'', leading to an incorrect container.

An immediate solution to this issue is to concatenate the adjacent instructions. However, this may lead to conflicting information as the instructions express different actions. (It can seen in the second row of Figure~\ref{fig:text_concat}).
Therefore, we propose leveraging LLMs' conversational understanding capabilities to re-caption the instruction into detailed input texts. 

Since the text-to-image models take the descriptive image captions as the input, we prompt the LLM to predict the state of the scene given each instruction. The process is illustrated in that ~\ref{fig:overview}, where the states $S = (s_0, s_1, ...s_N)$ are predicted  given a sequential set of $N$ instructions $A = (a_0, a_1, ...a_N)$.
We prompt the LLM by turns so that the model predicts the state of scene $s_i$ given the current instruction and previous instructions and states.
Therefore, the state $s_i$ contains the scene context after the first $i$ instructional steps, which we combine with the next instruction $a_{i+1}$ as the input prompts to the text-to-image model:
\begin{equation}
    p_i =
\begin{cases} 
a_i + s_{i-1} & \text{if } i > 0 \\
a_i & \text{otherwise}
\end{cases}
\end{equation}

We find that this re-captioning method makes individual text prompts contain the necessary information that is revealed and should be maintained in the previous steps. As a result, the text-to-image model can achieve better continuity and context accuracy in the generated visual instruction sequence.

\subsection{Dynamic consistent image generation}
Our re-captioning technique ensures the description continuity in text prompts, while we still need to generate consistent content shared across instructions. 
Unlike previous methods like video or story generation, which require absolute consistency on the entire scene or the object region, instructional generation requires a more nuanced consistency. 
As discussed in the previous sections, some visual instruction steps may demand maintaining object consistency, while others may involve significant transformations of the objects or background according to the procedure. To this end, we propose an adaptive KV-sharing method based on local region and state similarity constraints.

\begin{figure}[t]
    \centering
    \includegraphics[width=0.5\textwidth]{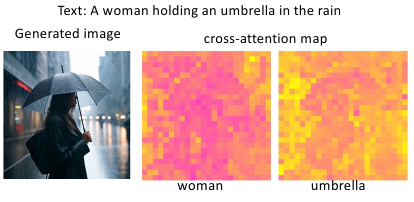}
    \caption{
    \textbf{Cross-attention map of Stable Cascade.} 
    We visualize the corss-attention maps in stage C of Stable Cascade model. It is found that the attention maps are noisy and fail to accurately delineate the specific regions of the main objects: woman and umbrella.
    }
    \label{fig:cross_attn}
\end{figure}

\myparagraph{KV sharing with local region constraint.} 
Similar to Consistory~\cite{consistory}, visual instruction generation only cares about consistency within certain regions. We identify these consistent objects in the instructions during our in-content planning phase. 
With our re-captioned instructions, the images generated realize most of the information about the instructions, like the layout, except for the object consistency. Therefore, we propose to use off-the-shelf segmentation models to generate the object masks~\cite{liu2023grounding,ren2024grounded}. We also explored using cross-attention maps to produce the mask, as in Consistory. 
However, the link between text and image can be weak and noisy in the state-of-the-art text-to-image models~\cite{stablecascade} as shown in Figure~\ref{fig:cross_attn}. Instead, we show that existing segmentation models work well with generated images and are not overfitted to a single model architecture.

\myparagraph{KV sharing with state similarity constraint.} To manage the dynamic scenario of generating consistent objects with variations, we continuously generalize the KV sharing. 
Instead of only sharing or not sharing KV features among different regions, we regulate the scaling of sharing with a state similarity matrix.
This matrix controls the degree of similarity between each pair of instructional steps. By adjusting its values, one can tailor the visual output to highlight consistency or emphasize transformation, depending on the instructions. 
To automatically infer this matrix from the instructions, we provide thoughtful instruction and in-context examples to LLMs.
Let $W \in [0, 1]^{N \times N}$ be the state similarity matrix, where $N$ is the number of steps. $W_{i,j}$ represents the overall similarity between the $i$-th and the $j$-th steps.

We inflate $W_{i,j}$ into a matrix $S_j$, which is used for computing the attention matrix of the $i$-th image. We use the matrices derived from the similarity matrix to regularize the attention maps in
Equation \ref{eq:attention_maps} is as follows:

\begin{equation}
S_i^+ = [S_1 \dots, S_{i-1} \oplus 1 \oplus S_{i+1}, \dots, S_N]
\end{equation}

\begin{equation}
A_i^+ = \text{softmax} \left( \frac{Q_i K^{+T}}{\sqrt{d_k}} + \log (S_i^+)  + \log(M_i^+)  \right) \\
\in \mathbb{R}^{P \times N \cdot P}
\end{equation}

Here, $A_i^+$ represents the adjusted attention weights, where the flow of information in self-attention is scaled inversely by the values in $S_i^+$. 
When $S_j$ approaches zeros (indicating no similarity between steps $i$-th and $j$-th states), the sharing of information between those specific steps is nullified. The magnitude of information sharing between the $i$-th and $j$-th steps is substantial when the values of $S_j$ are high, encouraging more consistent regions to be generated. This method allows us to precisely control the trade-off between consistency and variation across different steps in the visual instruction generation process, ensuring that each generated step is appropriately aligned with the instruction and the other images in the sequence.

\section{Experiments}
\label{sec:experiment}

\subsection{Experiment setup}
\label{sec:exp_detail}
In this section, we evaluate our method for generating visual illustrations of textual instructions. 
We propose a framework to leverage the vision language models (VLMs) for the evaluation. 
We also perform ablation studies on our individual designs.

\myparagraph{Evaluation}
Following the previous studies, we utilize CLIP-Score ~\cite{clip} to measure the text-image alignment, and Dreamsim ~\cite{dreamsim} and L2\_Dinov2 ~\cite{caron2021emerging} to evaluate the consistency.
However, the tasks in the instructions span a wide range of categories. 
Each of these may have a distinct goal beyond simple object consistency. 
Understanding the quality of the visual illustration demands nuanced reasoning---for instance. 
At the same time, some scenarios necessitate maintaining object consistency, and others require deliberate changes in the object's state, such as cutting or decorating. 
Therefore, conventional metrics often do not evaluate such a complex task. 
Instead, we turn to recent VLMs like GPT-4V and Gemini-Pro 1.5, which show great visual understanding and reasoning capacity.
We mainly evaluate in the following aspects:
\begin{enumerate}
    \item \textbf{Textual Alignment} measures how well the visual content matches the textual instructions.
    \item \textbf{Continuity} evaluates the transition process in a sequence of images or within elements of a single image.
    \item \textbf{Consistency} assesses whether the objects in the image remain the same throughout a sequence or within the context of the image.
    \item \textbf{Relevance} determines how focused the image is on the main object or theme as described in the input.
\end{enumerate}

To apply these metrics, we use carefully designed instructions and in-context examples to query vision-language models. 
We provide a pair of generated images, one by our method and the other by the baseline method, each time. The VLM is requested to pick the image better in the above aspects.

\myparagraph{Dataset}
To facilitate the study of visualizing textual instructions, we use GPT-4 to generate 200 goals and instructions for wide-ranging tasks, including cooking, gardening, and decorating. 
Each instruction contains 3 to 5 steps. 

\myparagraph{Implementation details.}
Our method includes two main components. 
We use GPT-4 API for in-context planning. We provide all the in-context examples and prompts in the appendix.
However, we note that users can achieve similar quality using ChatGPT3 or 4. 
For dynamic, consistent image generation, we use Stable Cascade ~\cite{stablecascade} as our text-to-image model, which has three stages: A, B, and C. 
Based on the report and exploration, we find that only stage C forms the image based on the text condition (stage B uses text as a condition, but in this stage, the text condition barely affects the generation results). 
Thus, we apply our adaptive KV sharing method in stage C. Specifically, we apply it in the first 15 steps, which total 20.

\subsection{Quantitative results}
\begin{figure}[t]
    \centering
    \includegraphics[width=0.5\textwidth]{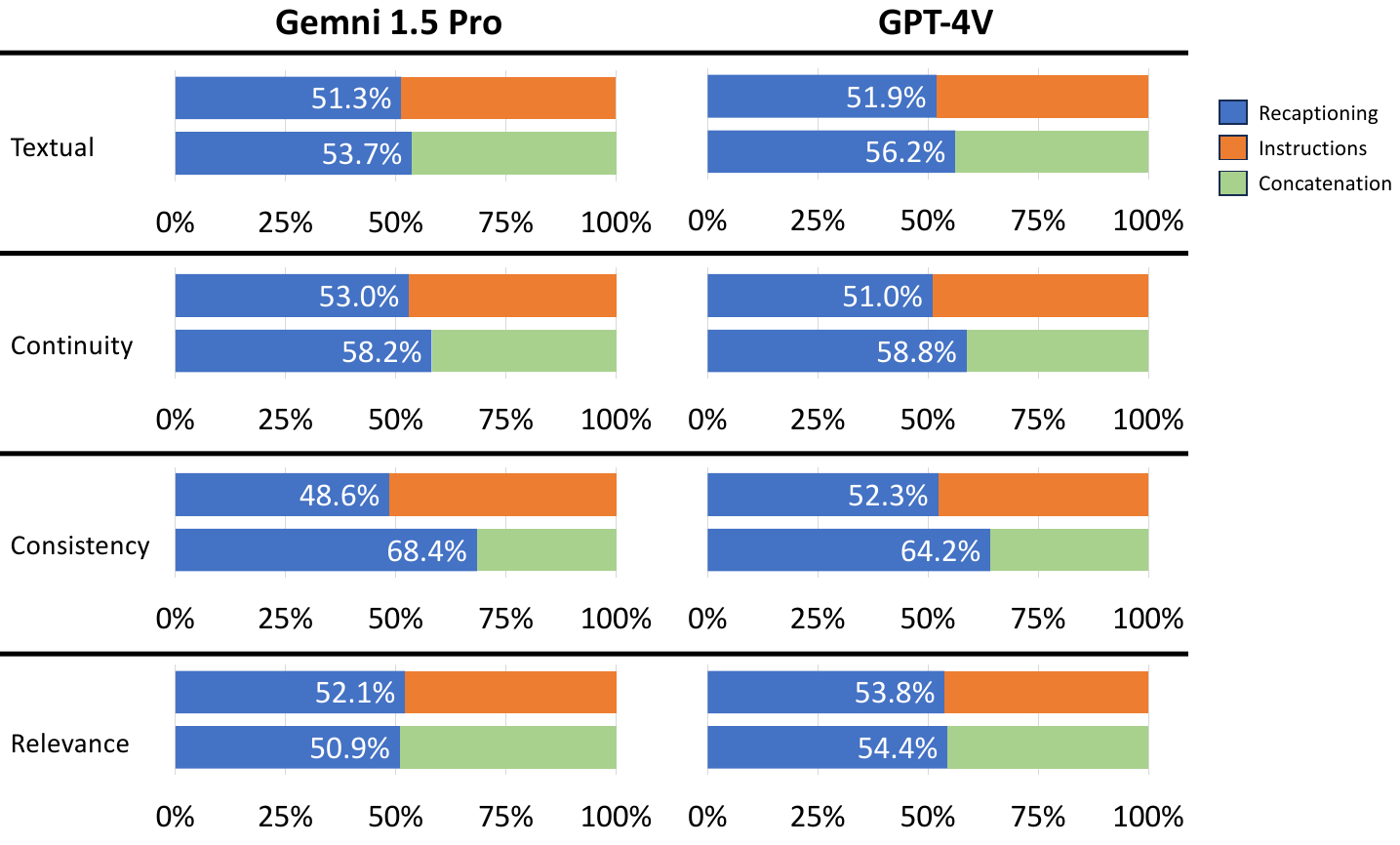}
    \caption{Evaluation of different design choices of text prompts using LLM, including Gemini and GPT-4. Among different evaluation aspects, including text alignment, continuity, consistency, and relevance, our choice of concatenating action and state beats using action only or concatenating with previous actions.}
    \label{fig:text_chart}
\end{figure}

\begin{figure}[t]
    \centering
    \includegraphics[width=0.5\textwidth]{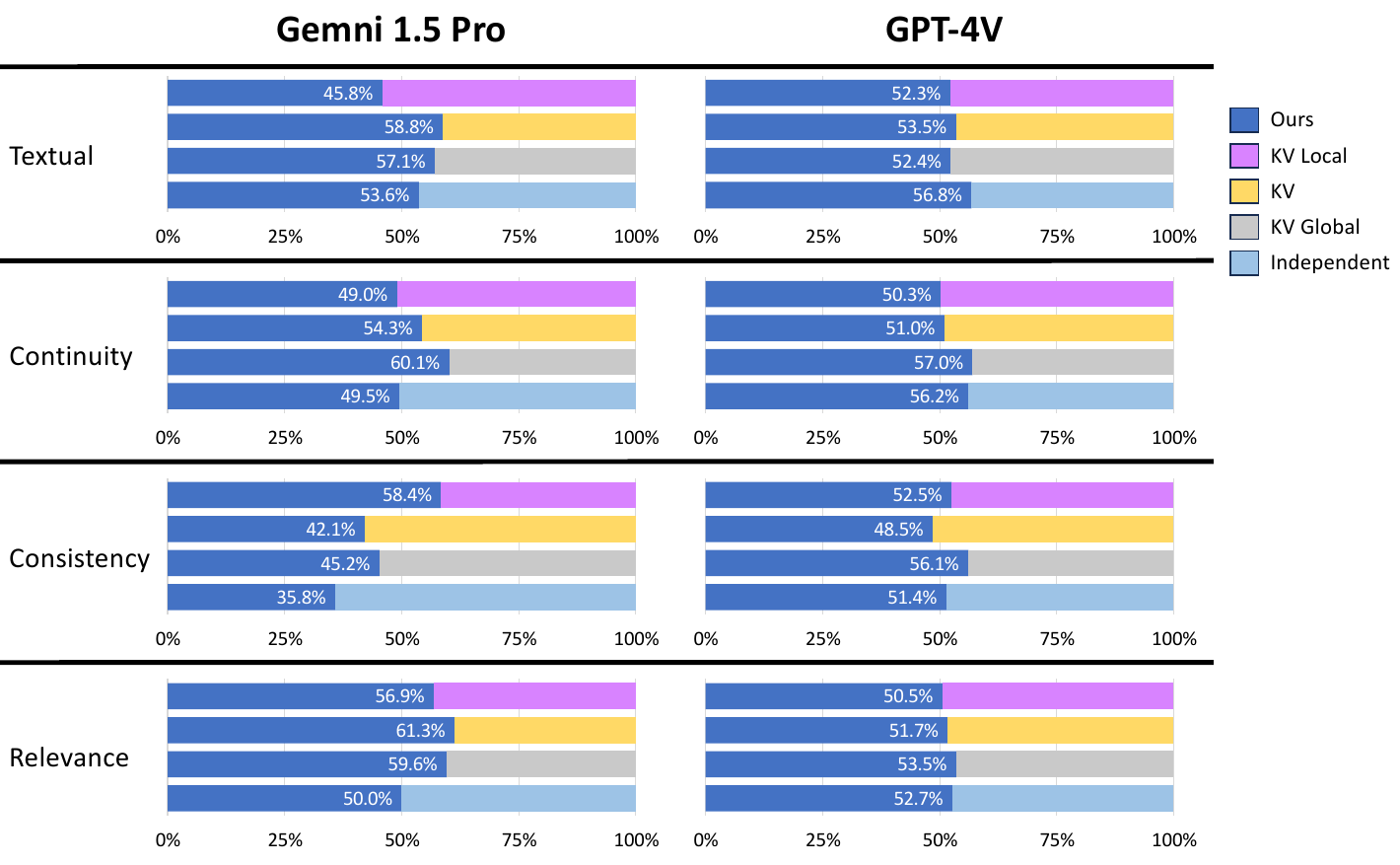}
    \caption{Evaluation of different consistency methods using LLM, including Gemini and GPT-4. Using both masks and weights achieves the best performances among all the choices overall.}
    \label{fig:chart_consistent}
\end{figure}

We evaluate our method and baseline approaches using vision-language models and show results in Figure~\ref{fig:text_chart} and Figure~\ref{fig:chart_consistent}. We quantitatively compare using different conditioning texts as input in Figure~\ref{fig:text_chart}, demonstrating that our re-captioning approach achieves overall better results than baseline methods relying solely on image generation instructions. 
Our method shows consistently improved behavior across all four aspects. We then quantitatively compare different feature-sharing methods in Figure ~\ref{fig:chart_consistent}. 
Our adaptive KV sharing methods improve almost all metrics using Gemini and GPT-4V.

We also show the quantitative results with traditional metrics in Table~\ref{tab:ablation_prompt} and~\ref{tab:ablation_image}. which demonstrates that our method is comparable with baselines. Specifically, as shown in Table~\ref{tab:ablation_image}, applying adaptive KV sharing leads to less identical results as desired while greatly improving the text alignment.

\begin{figure}[t]
    \centering
    \includegraphics[width=0.5\textwidth]{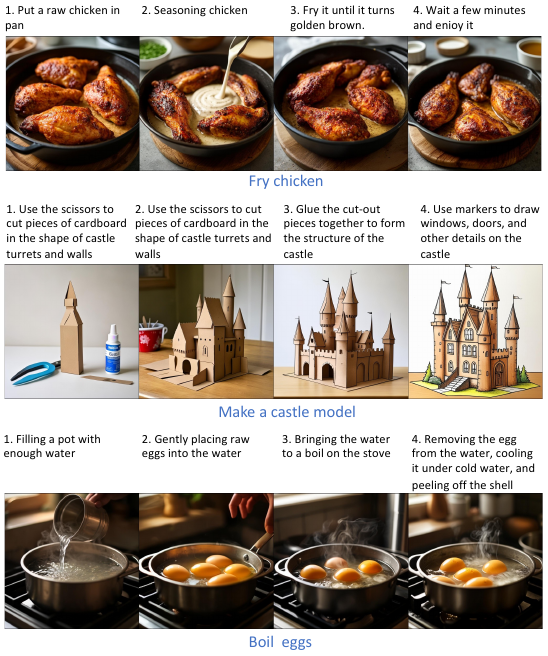}
    \caption{\tb{Failures cases.}
    Due to the limited capability of the text-to-image model, it cannot generate raw chicken or raw eggs.}
    \label{fig:fail}
\end{figure}

\begin{table}[t]
\centering
\setlength{\tabcolsep}{3pt}
\small
\caption{Ablation study of different design choices of the text prompts. 
Using instructions only provides the best text image alignment while concatenating with previous instructions or states improves the coherency.}
\label{tab:llm}
\begin{tabu} to 0.45\textwidth {X[1.6,l]X[1.2,c]X[1.2,c]X[1.3,c]}
\toprule
{Input texts} & {Clip-score} $\uparrow$ & {Dreamsim}$\downarrow$ & {L2-Dinov2}$\downarrow$ \\ 
\midrule
Instructions & \textbf{0.6980} & 0.0.4829 & 47.7013 \\
Concatenation & 0.4559 & \textbf{0.3433} & \textbf{38.8717} \\ 
Re-captioning & 0.5138 &0.3797 & 41.6487 \\ 
\bottomrule
\end{tabu}
\label{tab:ablation_prompt}
\end{table}

\begin{table}[t]
\setlength{\tabcolsep}{3pt}
\small
\centering
\caption{Quantitative evaluation of our methods to achieve object coherency. Our method, which uses both local region mask and global similarity constraints, greatly improves the text alignment for its flexibility in realizing the variation of object states across steps, which is shown as an increase in similarity measurement.}
\label{tab:llm}
\begin{tabu} to 0.45\textwidth {X[0.5,l]X[0.5,l]X[1.2,c]X[1.2,c]X[1.3,c]}
\toprule
{Global} & {Local} & {Clip-score} $\uparrow$ & {Dreamsim}$\downarrow$ & {L2-Dinov2} $\downarrow$ \\ 
\midrule
\xmark & \xmark &  0.4929 &\textbf{0.3638}   & \textbf{40.5037} \\
\cmark & \xmark & 0.5358 & 0.3792  &41.5642\\ 
\xmark & \cmark &0.5138 & 0.3799 & 41.6487 \\ 
\cmark & \cmark & \textbf{0.6708} & 0.4377 &45.6434\\ 
\bottomrule
\end{tabu}
\label{tab:ablation_image}
\end{table}

\label{sec:exp_quan}
\subsection{Qualitative results}

\begin{figure}[h]
     \begin{subfigure}[t]{0.01\textwidth}
        \vspace{-10.7cm}
        \rotatebox[origin=c]{90}{
            \small{$\overbrace{\hspace{2.cm}}_{\substack{\vspace{-6.0mm}\\\colorbox{white}{~~Ours~~}}}$ 
            $\overbrace{\hspace{2.cm}}_{\substack{\vspace{-6.0mm}\\\colorbox{white}{~~KV  global~~}}}$ 
             $\overbrace{\hspace{1.9cm}}_{\substack{\vspace{-6.0mm}\\\colorbox{white}{~~KV local~~}}}$  
            $\overbrace{\hspace{1.9cm}}_{\substack{\vspace{-6.0mm}\\\colorbox{white}{~~KV~~}}}$
            $\overbrace{\hspace{1.9cm}}_{\substack{\vspace{-6.0mm}\\\colorbox{white}{~~Independent~~}}}$
            }
   
        }
    \end{subfigure}
    \begin{subfigure}[b]{0.47\textwidth}
        \centering
        \includegraphics[width=1.\textwidth]{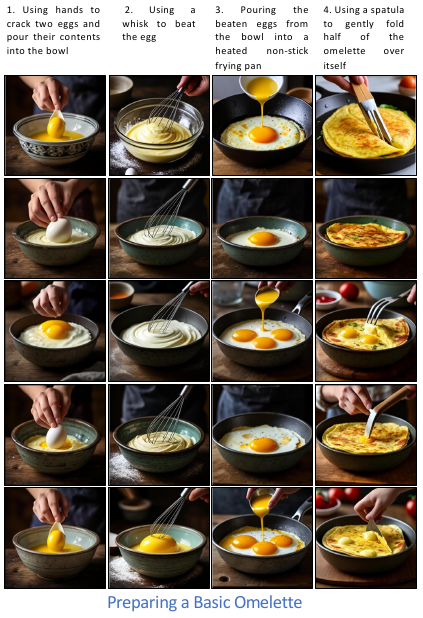}
        \vskip -0.2cm
    \end{subfigure}
    \vspace{-0.2cm}
    
    \caption{
    \tb{Visual comparisons of consistency across steps.}
    Here, we use the proposed prompting approach and focus on validating the various ways of sharing keys and values in attention layers for consistent image generation. 
    All methods take the same input text prompt.
    \textbf{KV}: sharing across early steps; 
    \textbf{KV local}: sharing controlled by masks; 
    \textbf{KV global}: sharing controlled by the proposed state similarity matrix; 
    \textbf{Ours}: sharing controlled by both masks and the state similarity matrix.
    For example, in the omelet, na\"lively sharing the key and value across steps cause \emph{content leaking} (e.g., the pan in steps 3 and 4 looks like the bowl in steps 1 and 2).
    With weight control, the method can maintain consistency, avoiding leaking feature while respecting the action. 
    Adding mask control helps improve textual alignment. (e.g, the actions are more align with text)
    }
    \label{fig:consistency}
    \vspace{-1em}
\end{figure}

\begin{figure*}[h]
     \begin{subfigure}[t]{0.02\textwidth}
        \vspace{-6.6cm}
        \small{
        \rotatebox[origin=c]{90}{
            \colorbox{white}{~~Re-captioning~~}
              \colorbox{white}{~~Concatenation~~}
               \colorbox{white}{~~Instruction~~}

        }
        }
    \end{subfigure}
    \begin{subfigure}[b]{0.97\textwidth}
        \centering
        \includegraphics[width=1.\textwidth]{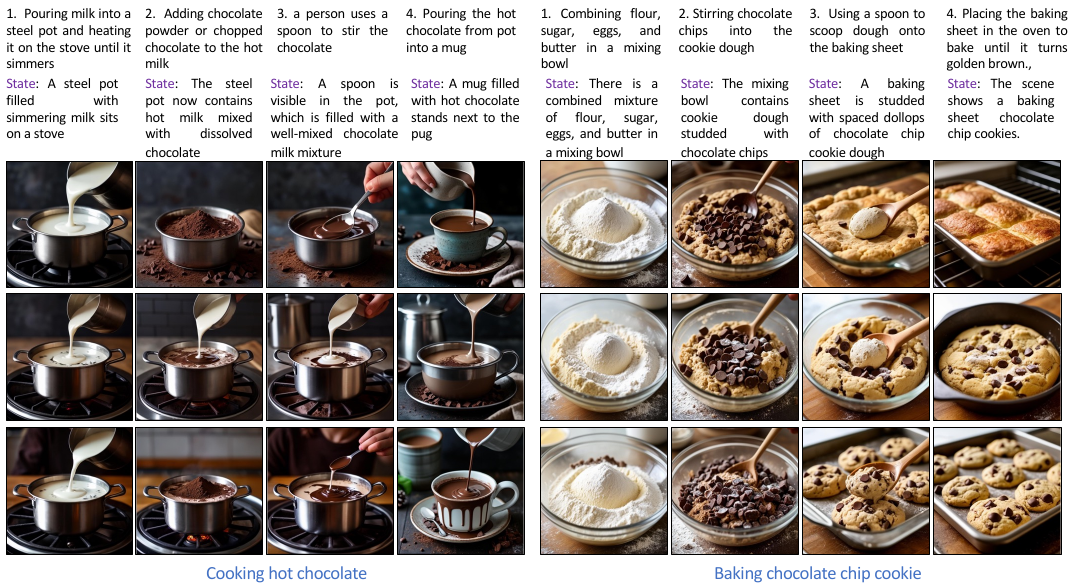}
        \vskip -0.2cm
    \end{subfigure}
    \vspace{-0.2cm}
    \caption{
    \tb{Visual comparisons with concatenating steps.}
    \textbf{Ours}: concatenating current instruction and inferred states; \textbf{Concatenation}: concatenating current and previous instructions; \textbf{Instruction}: using only current instruction
    }
    \label{fig:text_concat}
    \vspace{-1em}
\end{figure*}

\begin{figure*}[h]
     \begin{subfigure}[t]{0.02\textwidth}
        \vspace{-6cm}
        \rotatebox[origin=c]{90}{
            $\overbrace{\hspace{2.6cm}}_{\substack{\vspace{-7.0mm}\\\colorbox{white}{~~Ours~~}}}$ 
            $\overbrace{\hspace{2.6cm}}_{\substack{\vspace{-7.0mm}\\\colorbox{white}{~~Genhowto~~}}}$

        }
    \end{subfigure}
    \begin{subfigure}[b]{0.97\textwidth}
        \centering
        \includegraphics[width=1.\textwidth]{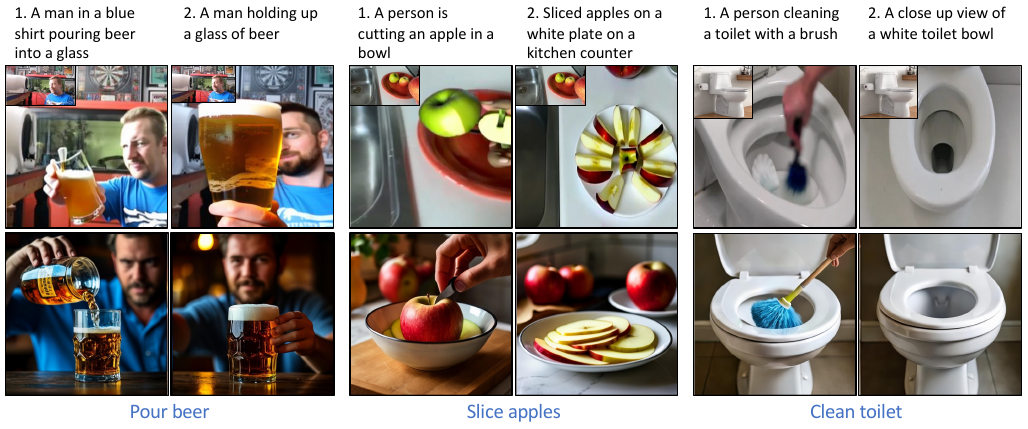}
        \vskip -0.2cm
    \end{subfigure}
    \vspace{-0.2cm}
    
    \caption{
   \tb{Visual comparisons with Genhowto ~\cite{genhowto}}. Comparison of our method with the GenHowTo method on instructions from their paper. The small images in the top left of the GenHowTo images are the real images used as inputs for their method. Our free-training method is compatible with their pretrained model and produces visually more pleasing results without requiring a real image as input.
    }
    \label{fig:genhowto}
    \vspace{-1em}
\end{figure*}

Figure ~\ref{fig:text_concat} illustrates the impact of various design choices for prompts on the coherence and alignment of the generated instructions. When only the action from a single step is used as input, the generated images lack coherence due to unawareness of the context. 
For instance, in ``cooking Hot Chocolate", the pot is only mentioned in step 1.  
As a result, steps 2 and 3 do not know that the pot is in the stove and contains the hot milk. Therefore, the model fails to maintain the context; the pot is not in the stove and does not contain the milk anymore in steps 2 and 3. 
In scenarios where current and previous actions are concatenated, the model can better understand the context. 
However, such models tend to prioritize the most recent action at the cost of accurately generating earlier ones. Concatenating action and state significantly enhances context understanding, leading to more accurate image generation based on prior steps. For instance, this method ensures the generated image accurately places the pot on the stove in the case of a pot and cooking. 
In another example involving cookies, the approach allows for generating images of cookies with chocolate chips, ensuring the action aligns closely with the input text.

In Figure~\ref{fig:consistency}, the stable cascade model, when generating images independently, fails to ensure consistency across different steps. Using basic KV sharing makes images in different steps globally similar. However, errors occur, such as a pan being misinterpreted as a bowl and the actions not aligning with the input text well. 
Applying a local region mask addresses some issues of action omission, though it still misrepresents a bowl as a pan. Regularizing KV sharing with state similarity resolves the problem of missing attributes between steps, yet it inadequately captures the action, as seen in the poor depiction of pouring eggs. 
Ours combines both methods and effectively solves the issues of missing actions and object misrepresentation across different steps.

In Figure ~\ref{fig:genhowto}, we compare our method with GenHowTo~\cite{genhowto}, a pretrained model for generating actions and states from real images. We use prompts from the GenHowTo test dataset, which belongs to the same categories as their training data. Note that our method is training-free and does not require a real input image as input. 
Ours produces comparable results regarding textual alignment, consistency, and even better visual quality. This is because our approach leverages a state-of-the-art text-to-image model. Unlike GenHowTo ~\cite{genhowto}, which requires re-training the model for each new base text-to-image model and may compromise the generation quality due to the training data quality.

\section{Failure cases and discussion}
We shown several failure cases  within our model in Figure ~\ref{fig:fail}, where it did not accurately generate the stated objects and attributes from the instructions, thereby misleading the process. For example, in a step described as frying raw chicken, the model erroneously generated an image of cooked chicken. Similarly, in another instance involving boiling a raw egg, the output also deviated from the specified raw state. Additionally, the model exhibited a bias towards rendering castles in a painting style, which led to inconsistencies in the style of generation across different tasks. We believe that as text-to-image models improve in the future, our method will greatly benefit from reduced limitations inherent in current text-to-image models.

\section{Conclusion}
\label{sec:conclucion}

In this paper, we tackle the problem of generating static visual illustrations from textual instructions by leveraging pretrained diffusion models, enabling high-quality generation without expensive fine-tuning. We propose a framework to address the unique challenges of maintaining object consistency across instructional steps while managing the variability of objects that change across states. Our extensive evaluations demonstrated that our method outperforms baseline models in both consistency and accuracy.

{\small
\bibliographystyle{ieeenat_fullname}
\bibliography{sample-base}
}

\appendix

\section{Appendix}
\label{sec:appendix}

\subsection{Additional discussion and details about large language models}
\label{sec:gpt-4}
\myparagraph{Full prompt for GPT-4}
Our complete prompt for GPT-4 includes three main components:

\begin{itemize}
    \item \textbf{Instruction:} This specifies the task and defines the output format, helping GPT-4 perform effectively in layout generation tasks.

    \item \textbf{In-context exemplars:} These enhance the model’s capability for the task by providing multiple examples. These examples help the model understand the context better and produce the desired bounding boxes and corresponding labels.

    \item \textbf{User prompt:} This is appended to the instruction and supporting examples. The model then continues the conversation based on the user prompt and provides the layout in the specified format.
\end{itemize}

When users provide a prompt (user prompt), it is combined with the predefined text to create a complete prompt as shown in Table ~\ref{tab:chatgpt}. The GPT-4 API then processes this complete prompt and returns the information about each steps, similarity matrix and the main objects in each step.

\myparagraph{Metrices using Multimodel Gemini and GPT-4V}
We evaluate our genererated visual instruction using Gemini15-pro and gpt-4V to evaluate four aspects, we use the instruction sjown in Figure ~\ref{fig:metrics} to guide the multi-model to assess our visual instructions. We shuffle the order of the two methods which are compared to avoid the multi-model bias toward the order.

\begin{table*}
\centering
    \caption{The full prompt for gpt4 api to generate instructions.  }
    \label{tab:chatgpt}
    \tabulinesep=4pt
    \begin{tabu} to 1.\textwidth {@{}X[1.9,l]X[9,l]@{}}
    \toprule
    \textbf{Role} & \textbf{Content} \\
    \midrule
      \textbf{Instruction}& System:{ "You are ChatGPT-4, act like visual and instructional experts, generate step-by-step how to do something. each step include the action to indicate how people interact with objecs, and state to show state of objects after finish this action. And relation matrix is the correlation of one step with others in visual. object field indicate the objects in each step similar with privious step in some extends: similar(total similar), shape similar(only similar shape), texture similar( transform shape, only same texture)"}\\
    \midrule
     \multirow{2}{*}{\textbf{In-context examples}} & User: {"The instruction on decorating a cake in 2 steps."}\\
     & Assistant: [examplers in Figure ~\ref{fig:json_dict}] \\
     \midrule
    \textbf{User prompt} & User : "The instruction on" + [user prompt]\\
    \bottomrule
    \end{tabu}
\end{table*}

\begin{figure}[ht]
\centering

\lstset{
  basicstyle=\ttfamily,
  breaklines=true,
  frame=single,
  backgroundcolor=\color{white},
  showstringspaces=false
}
\begin{lstlisting}
{
    "goal": "Decorating a Cake",
    "steps": [
        {
            "step": "Setting the Cake on a Platter",
            "object": [["cake", "new"], ["platter", "new"]],
            "action": "Set the baked cake on a platter.",
            "state_of_main_object": "A baked cake on the platter."
        },
        {
            "step": "Applying Icing",
            "object": [["cake", "similar shape", 1], ["spoon", "new"]],
            "action": "Person using a spoon to place some icing on the top of the cake.",
            "state_of_main_object": "The cake covered by icing."
        }
    ],
    "relation": [
        [1.0 , 0.5, 0.4, 0.3],
        [0.9, 1.0 , 0.5, 0.4],
        [0.8, 0.9, 1.0, 0.4],
        [0.7, 0.8, 0.9, 1.0 ]
    ]
}
\end{lstlisting}
\caption{In-context examplers for full prompts}
\label{fig:json_dict}
\end{figure}

 \clearpage
 
\begin{figure*}[h]
\centering

\begin{minipage}{\textwidth}
\lstset{
  basicstyle=\ttfamily,
  breaklines=true,
  frame=single,
  backgroundcolor=\color{white},
  showstringspaces=false
}
\begin{lstlisting}

   
Our task here is to compare visual step-by-step instructions, generated from the same step-by-step textual instruction. We want to decide which one is better according to the provided criteria.
# Instruction
1. Text prompt and Asset Alignment: Focus on whether the key elements mentioned in the text are clearly visible and identifiable in the image. The visual is good if all key elements are clearly depicted and easily identifiable.
2. Continuity: This measures how well the image captures the progression from the previous step(s), maintaining context and demonstrating the changes or actions described in the current step. The visual is good if the image effectively shows the progression from previous steps and integrates new elements/actions as described in the current step.
3. Consistency: Evaluates whether the same objects are used consistently across all images in a way that reflects their continued presence and role as described in the text. This is particularly important for objects that are central to the action or instructions. For example, a pot in first step should look like the pot mentioned other step, even it can be in different views.
4. Relevance: Assesses whether the visual focuses on the most critical aspect of the step as described in the text. The visual is good if the visual focuses precisely on the primary action or element described in the step.
Take a really close look at each of the multi-image instructions for the corresponding textual instruction before providing your answer.
When evaluating these aspects, focus on one of them at a time.
Try to make independent decisions between these criteria.
# Output format
To provide an answer, please provide a short analysis for each of the abovementioned evaluation criteria. The analysis should be very concise and accurate.
For each of the criteria, you need to make a decision using these options:
1. The first row visual is better;
2. The second row visual is better;
... or Cannot decide.
IMPORTANT: PLEASE USE THE 'Cannot decide' OPTION SPARSELY.
Then, in the last row, summarize your final decision by <option for criterion 1> <option for criterion 2> <option for criterion 3> <option for criterion 4>.
# Example

Analysis:
1. Text prompt and Asset Alignment: The first one ...; The second one ...; The first/second/third/... one is better or cannot decide.
2. Continuity: The first one ...; The second one ...; The first/second/third/... one is better or cannot decide.
3. Consistency: The first one ...; The second one ...; The first/second/third/... one is better or cannot decide.
4. Relevance: The first one ...; The second one ...; The first/second/third/... one is better or cannot decide.
Final answer:
x, x, x ,x (e.g., 1, Cannot decide, 3, 1/ 2, Cannot decide,5, 1 / 1, 3, 2,4)

\end{lstlisting}
\end{minipage}
\caption{Instruction for Gemini and GPT-4V to asses the visual instruction}
\label{fig:metrics}
\end{figure*}

\end{document}

\maketitle
\input{sections/sup_compare}
\input{sections/sup_grounded_implementation}
\input{sections/sup_layout}
\input{sections/sup_quantitative}
\input{sections/sup_instruct}
\input{sections/sup_ablation}